\documentclass[conference]{IEEEtran}

\usepackage{cite}
\usepackage{amsmath,amssymb,amsfonts}
\usepackage{mathabx}
\usepackage{algorithmic}
\usepackage{graphicx}
\usepackage{textcomp}
\usepackage{xcolor}
\def\BibTeX{{\rm B\kern-.05em{\sc i\kern-.025em b}\kern-.08em
    T\kern-.1667em\lower.7ex\hbox{E}\kern-.125emX}}

\newcommand{\E}[1]{E_{P(e|z)}\left[#1\right]}

\newcommand{\set}[1]{\left[#1\right]}

\newcommand{\eqnref}[1]{(\ref{eqn:#1})}

\newcommand{\eqnlabel}[1]{\label{eqn:#1}}

\newcommand     {\paren}[1]{\left(#1\right)}
\newcommand{\abs}[1]{\left|#1\right|}
\newcommand{\curlb}[1]{\left\{#1\right\}}

\begin{document}

\title{\Large VQalAttent: a Transparent Speech Generation Pipeline based on Transformer-learned VQ-VAE Latent Space
}
\author{
\IEEEauthorblockN{\normalsize Armani Rodriguez, Silvija Kokalj-Filipovic}
\IEEEauthorblockA{\small Rowan University\\
\small\em rodrig52@students.rowan.edu, kokaljfilipovic@rowan.edu}}

\maketitle
\begin{abstract}
Generating high-quality speech efficiently remains a key challenge for generative models in speech synthesis.  This paper introduces VQalAttent, a lightweight model designed to generate fake speech with tunable performance and interpretability. Leveraging the AudioMNIST dataset, consisting of human utterances of decimal digits (0-9), our method employs a two-step architecture: first, a scalable vector quantized autoencoder (VQ-VAE) that compresses audio spectrograms into discrete latent representations, and second, a decoder-only transformer that learns the probability model of these latents. Trained transformer generates similar latent sequences, convertible to audio spectrograms by the VQ-VAE decoder, from which we generate fake utterances.  Interpreting statistical and perceptual quality of the fakes, depending on the dimension and the extrinsic information of the latent space, enables guided improvements in larger,  commercial generative models. As a valuable tool for understanding and refining audio synthesis, our results demonstrate VQalAttent’s capacity to generate intelligible speech samples with limited computational resources, while the modularity and transparency of the training pipeline helps easily correlate the  analytics with modular modifications, hence providing insights for the more complex models. 
\end{abstract}

\vspace{-2mm}
\section{Introduction}
State-of-the-art generative neural speech synthesis and  speech  representation learning \cite{mohamed2022self} are usually influenced by the latest AI research in computer vision, which is then adapted and extended leveraging classical speech processing theory and practice. This has been the pattern ever since the first neural synthesis by Wavenet \cite{oord2016wavenet}.  However, it is the availability of massive computational
resources that allowed  breakthroughs such as
WaveNet  or Tacotron 2 \cite{shen2018natural}. Similar influences exist from Natural Language Processing (NLP)  \cite{wu2024towards}, which brought transformers and large language models (LLMs) into the speech synthesis scene  \cite{HuangSpSynth, defossez2022high}.  Majority of these neural synthesis models are focused on text-to-speech (\cite{shen2018natural},\cite{chen2024valle}, \cite{wu2024towards} and the references therein).

In this paper we introduce and analyze a simple model composed of a learned vector quantizer and a decoder-only transformer.  Although of limited scope due to a limited-complexity  training dataset, this model allows us to analyze the effects of synthesizing speech in discrete domain, which is enabled by the learned quantization of the latent space of raw acoustic waveforms transformed to  spectrograms. Discrete domain synthesis has much lower computational complexity, which creates an opportunity for experimentation and analysis despite the lack of massive computational resources.

{\bf Prior Work.} 
WaveNet, a generative neural 
network introduced by Oord et. al. \cite{oord2016wavenet} utilizes dilated convolutions to generate raw waveforms 
autoregressively. The architecture was 
based off another autoregressive 
generative model, PixelCNN, which 
generated images pixel by pixel.
WaveNet model found success with
generation of audio 
data, including speech in various 
languages and music. While it can generate high 
fidelity audio and comprehensible 
speech, the autoregressive method makes WaveNet extremely 
slow for both training and inference. 
This is making it inappropriate for use cases 
that require quick synthesis of audio.
Donahue et. al. \cite{donahue2018adversarial} introduced 
WaveGAN, a generative adversarial 
network (GAN) for raw waveform data. In the paper, Donahue et. 
al. compared the results of WaveGAN
against the performance of another GAN model, called 
SpecGAN. SpecGAN utilizes spectrogram 
representations of waveforms which are 
more fitting for machine learning 
tasks than the waveform representation. 
The results were interesting, with the 
SpecGAN-generated samples producing more intelligible results but the 
WaveGAN generated samples being 
preferred by human judges. This was
likely due to the distortion resulting from
the fact that perceptually-informed spectrograms (Mel spectrograms) are not invertible without loss \cite{GriffinLim}. Similarly, we trained our speech synthesis using a dataset of spectrograms, and we observed the same audio artifacts when the synthesized data is played back. However, this is of no importance here, as the play-back quality could be improved by utilizing invertible transforms or by a post-hoc vocoder, while the statistical properties of the generated samples in the spectrogram domain showed desired similarity with the originals. 
HuBERT \cite{hsu2021hubert}, a very popular approach to learning self-supervised speech representations for speech recognition, generation, and compression, was one of the first models that used clusterization to discretize the space of speech signals. Other models  use AI discrete codecs to learn a well-discretized latent space and then train an autoregressive model in that discrete space \cite{mohamed2022self}. This alleviates the computational latency problem, inherent to  autoregressive approaches. There are multiple models that leverage the ability of VQ-VAE \cite{vqvae2017neural} to learn discrete latent representations \cite{dhariwal2020jukebox}, which are particularly useful for capturing the complex structures present in audio data. 

Hence, the type of approach that we are using is not novel, but other models that leverage VQ-VAE for discretization are overly complex, with diverse components and multi-modal loss functions, which makes the analysis very difficult. For example, \cite{defossez2022high} utilizes multiple quantization codebooks for variable bandwidth in the end-to-end training of the model pipeline  which consists of a transformer, a GAN architecture, an arithmetic encoder and several loss components, including multiple reconstruction losses and an adversarial loss mechanized by the GAN. As the speech synthesis in discrete domain is fairly new, we believe that VQalAttent, the model with a straightforward architecture proposed here, will contribute to better understanding of this approach. 
The main contribution of this architecture is the model transparency, allowing for a scalable model adaptation, followed by the comparative analysis based both on tracking the shift in learned probability distribution and on assessing classification accuracy. Adaptations demonstrated here include dimensionality of discrete representations (compression ratio),  codebook initialization, conditional generation with the condition prompt embedded in the discrete representation sequence, and the choice of the transformer model.
\section{System Model} \label{sec:sysmodel}
Our methodology involves two 
models.  We use a vector-quantizing auto-encoder (VQ-AE), which is typically referred to as VQ-VAE, despite the fact that it is not trained using variational inference. For this reason, we will also use the acronym VQ-VAE, since the architecture that we are using allows for variational inference using a Kullback-Liebler (KL) distance between the uniform probability mass function of discretized data and the learned posterior probability mass function.  We use VQ-VAE to learn discrete latent representations of the audio recordings in AudioMNIST,  converted to spectrograms. The training involves choosing a discretization model that has the most perceptible reconstructions since the 
quality of the reconstructions reflect the quality of the learned representation.
Using these learned discrete latents,  we train an autoregressive  model to learn the prior over the  latents representing the training dataset. The particular model that we use is a decoder-only-transformer with $X=12$ self-attention blocks, each with $Y=8$ heads.  Through autoregressive sampling of the trained transformer, we generate deep fakes. 
In the absence of human judges to 
score the intelligibility of our deep fakes, 
we will use the accuracy score of a deep classifier to determine how perceptible 
the reconstructed samples are (Table~\ref{tab:acc}). We will also measure the shift in learned probability distribution with respect to original data by characterizing the overlapping between the original and the fake manifolds using metrics such as fidelity (coverage) and diversity. 
\subsection{Vector Quantized Variational Autoencoder}
\begin{figure}[h]
\vspace{-2mm}
\centering
\hspace{-1mm}\includegraphics[width=0.46\textwidth, height = 4.2cm]{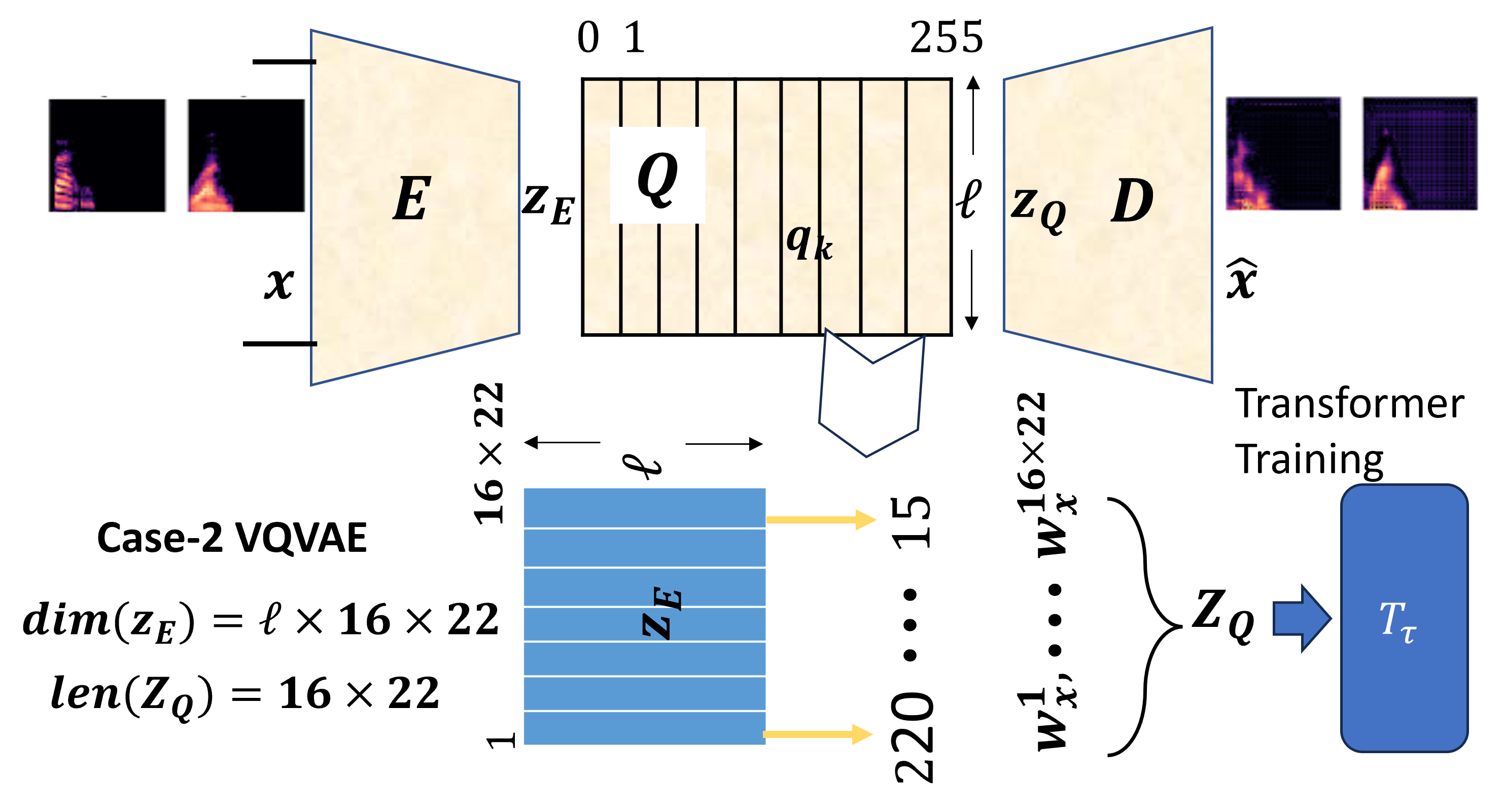}
\vspace{-2mm}
\caption{Architecture of VQVAE. The dimension and size of the codebook match the configuration of VQalAttent - Case 2. 
}\vspace{-3mm}
\label{fig:vqvaeglob}
\end{figure}

Vector Quantized Variational Autoencoder (VQ-VAE) \cite{vqvae2017neural} is an autoencoder which learns a discrete latent representation of the input data, in contrast to the continuous latent space learning in standard variational and non-variational autoencoders. The encoder, denoted $ E(x) $, transforms the input data $x$ into a sequence of latent vectors of length $\ell$. These latent vectors are then discretized by applying a quantization function $ \Phi $, which maps each latent vector to an element of a learnable codebook $ Q  = \{ e_i \mid 0 \leq i < N\}$ consisting of $N$ learnable embedding vectors of the same length ${\ell}$. The decoder block, denoted $ D(e) $, is tasked with reconstructing the original input $x$ from the mapped codebook vectors. While the most intuitive and most common choice for $ \Phi $ is to simply perform a nearest neighbor lookup, more advanced techniques such as stochastic codeword selection $\Phi_{SC}$ \cite{StochQ} or heuristics such as exponential moving averages (EMA) $\Phi_{EMA}$ \cite{vqvae2} can be employed instead. We utilized both. The 3 elements of the VQ-VAE architecture are shown in Fig.~\ref{fig:vqvaeglob} and the forward path through VQ-VAE is given by Equation~\eqnref{vqvaefwd} in which $z$ is the latent representation of $x$ at the output of $E$ while $z_q$ is its quantized version. We introduce another form of the quantized latent $z_q$, denoted in Fig.~\ref{fig:vqvaeglob} by $Z_Q,$ which replaces the codewords quantizing the $z'$s vectors with their indices in the codebook. We refer to those indices as tokens $Z_Q^j \in \curlb{0,\cdots N-1}, N=\abs{Q}=256.$  
\begin{align}
\nonumber z &= E(x, \theta_E) \\
\nonumber z_q &= \Phi(z, Q, \theta_Q) \\
    \hat{x} &= D(z_q, \theta_D) \eqnlabel{vqvaefwd}
\end{align}
More rigorously, the indexed (discrete) $Z_Q^j$ is obtained as
\begin{equation}
Z_Q^j = argmin_{0\leq i\le \abs{Q} }\E{\left\|z_j-e_i\right\|_2^2}, \eqnlabel{indexed}
\end{equation}
where the expectation $\E{}$ depends on the quantization algorithm $\Phi(z, Q, \theta_Q),$ $e_i$ is the $i$th codeword of the codebook $Q$ and $z_j$ is the $j$th slice (vector) of the latent tensor $z,$ of length $\ell$ equal to the length of $e_i.$
In Fig.~\ref{fig:vqvaeglob}, the token's index $j$ goes from $1$ to $16\times22,$ which is how many vectors of length $\ell=64$ the $z$ tensor consists of, for a specific compression rate (to be referred as the Case 2 VQ-VAE).
The architecture of the encoder is composed of 3 convolutional layers and 3 residual blocks, while the decoder has an equivalent symmetric architecture. The VQ-VAE, whose loss function is composed of a reconstruction loss $L_R = MSE(x,\hat{x})$ and a latent quantization loss  (commitment cost) $L_Q=MSE(z,z_q)$, is trained for 100 epochs. For $\Phi_{SC}$, we include a KL loss between $P(e_i|z)$ and a uniform prior. Note that the VQ-VAE trainable parameters are represented in Equation~\eqnref{vqvaefwd} as $\theta_E,$ $\theta_Q$ and $\theta_D,$ denoting  Encoder's weights, Codebook $Q$ trainable codewords and Decoder's weights, respectively.
\subsection{Decoder-Only Transformer}\label{subsec:donly}
\begin{figure}[h]
\vspace{-2mm}
\centering
\hspace{-1mm}\includegraphics[width=0.48\textwidth, height = 4.3cm]{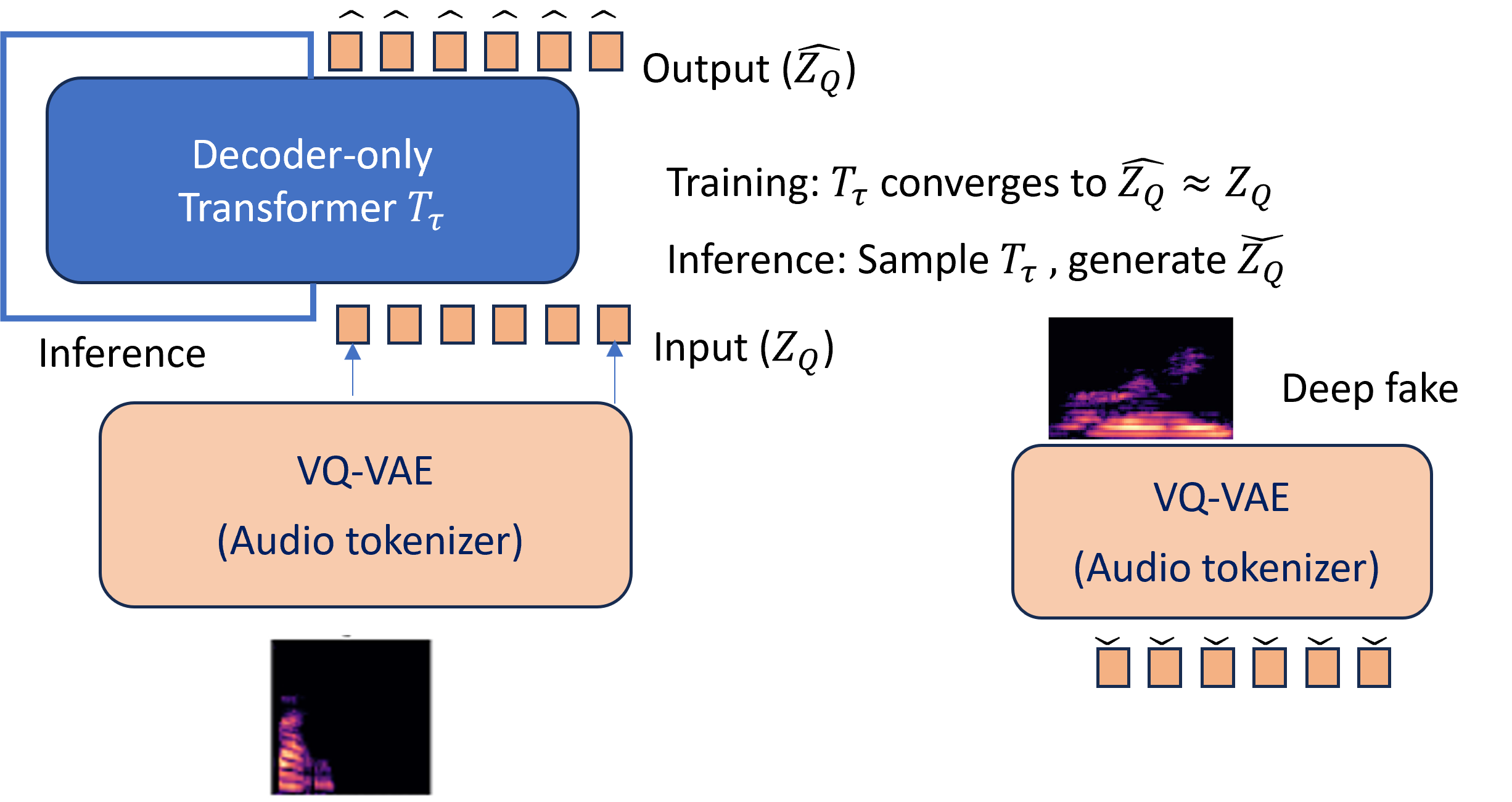}
\vspace{-2mm}
\caption{Architecture of VQalAttent and data flow in the training and inference phase: training latent sequences are denoted $Z_Q,$ training output of the transformer by $\widehat{Z_Q}$ and the autoregressively generated fake by $\widecheck{Z_Q}.$
} \vspace{-3mm}
\label{fig:vqattentglob}
\end{figure}
    The Decoder-only Transformer model \cite{deconly}, also known as the autoregressive Transformer, is a variation of the original encoder-decoder transformer model proposed by Vaswani et. al. \cite{vaswani2017attention}. This model omits the encoder stack, utilizing only the decoder stack to learn the causal attention structure in $Z_Q,$ and to perform autoregressive sequence generation, once trained. The model's self-attention mechanism enables the model to predict the next token in a sequence by attending to all previous tokens. It also allows for the use of cross-attention, but  does not specify how the cross-attending context is  obtained (as opposed to an encoder-decoder architecture where the cross-attention context is coming from the encoder).  As illustrated in Fig.~\ref{fig:vqattentglob}, during the training of the transformer, the masked indexed output of the VQ-VAE quantizer $Z_Q$ is compared with its estimate $\hat{Z}_Q$ at the output of the transformer using a cross-entropy loss. At the inference, we start from a random token, and continue generating subsequent tokens of a fake $\hat{Z}_Q$ auto-regressively. We convert $\hat{Z}_Q$ \eqnref{indexed} into its non-indexed version $\hat{z_q}$ and feed it into the VQ-VAE decoder to generate the spectrogram of the discrete latent deep fake produced by the transformer.
\section{Methodology}
In this section, we outline how our model is parameterized and trained, and how we design experiments to explore the following questions:
\begin{itemize}
    \item Can our model generate coherent, comprehensible human speech samples?
    \item Can the model be extended to conditionally generate speech based on content (e.g., digit labels) and speaker characteristics (e.g., gender)?
    \item How the model adaptation affects the above aspects?
\end{itemize}

\subsection{Data}
We use the AudioMNIST dataset, 
consisting of labeled, short audio 
samples of spoken numerals (0-9) \cite{audioMNIST}. 
The dataset uses a variety of speakers 
of different ethnicities and genders.

We opt for a spectrogram representation, a visualization of the signal frequency spectrum over time. Specifically, we will use the Mel-spectrogram. The Mel-spectrogram converts frequencies to Mel scale which is meant to approximate humans’ non-linear frequency response, i.e., our greater sensitivity to changes in low frequencies. Our complete data preprocessing pipeline is as follows:

\begin{enumerate}
    \item Resample audio recordings to 22,050 Hz. 
    \item Trim the leading and trailing 
silence, where 15dB under the peak 
amplitude is considered silence.
\item Fix the length of sampled recordings to 
a fixed duration by performing a time stretch.
    \item Apply the Mel-spectrogram 
transformation to obtain spectrograms as images of size $1 \times 64 \times 88$, with the number of gray pixels totaling $5,632$ per spectrogram.
\item Convert spectrogram values from linear-scale 
to decibels.
\item Apply min-max 
normalization.
\end{enumerate}
We refer to the training dataset composed  of such spectrograms as $X_{train}.$ Similarly, test dataset is referred to as $X_{test}.$
\subsection{Training}
We utilize a two-stage training procedure: first, we train the VQ-VAE model, followed by the training of the autoregressive transformer $T_{\tau}$ in Fig.~\ref{fig:vqattentglob}. 

As described in Section~\ref{sec:sysmodel}, in the initial training stage the VQ-VAE model will learn the discrete embeddings to achieve an efficient, discretized representation of the input data, which can be reconstructed by the VQ-VAE decoder into speech utterances perceptually resembling the originals. This is the non-generative part.  Such discrete ordinal data is well-suited for our transformer model $T_{\tau}$ with minimal preprocessing. The $T_{\tau}$ model is then trained on the latent sequences of tokens representing datapoints in our training set. We train 2 versions of this 2-stage pipeline: 1) {\em $VQVAE^{case 1}$}, which has longer token sequences due to a VQ-VAE encoder dimensioned to compress the spectrograms with compression ratio $CR=4$, from the original $1 \times 64 \times 88$, to $1 \times 32 \times 44 = 1408$ dimensions (ordinal tokens); 2) {\em $VQVAE^{case 2}$}, which produces shorter token sequences due to compression ratio $CR=16$, from the original $1 \times 64 \times 88$, to $1 \times 16 \times 22 = 352$ dimensions (ordinal tokens). 

Dimensioning of the VQ-VAE/ transformer pipeline is determined by the desired dimension of the latent tensor $Z_Q,$ which was guided by the following experiment: we performed a PCA decomposition of $X_{train},$ and counted the number of components $n_{99}$ whose explained variance constitute 99\% of the total variance. This number $n_{99}=1480,$ determined the dimension of $Z_Q$ in Case 1 (which is close, while easily implemented), while Case 2 was obtained by decreasing the second and third dimension of the encoder's output $z$ by half. Hence $VQVAE^{case 1}$ is expected to reconstruct the data with high fidelity if properly trained.

{\bf Learning the $Z_Q$ posterior.} In a Bayesian treatment we imply that latent variables govern the distribution of the data. Hence, looking at VQalAttent via a Bayesian model, we want to generate deep fakes by drawing latent variables $Z$ from a prior distribution 
$p(Z)$ and then relate them to the observations through the likelihood 
$p(x|Z)$  implemented by the VQVAE decoder. However, with the quantization step in VQ-VAE, we trained a discrete transform that maps the $Z$ latents into  discrete latents $Z_Q$. More precisely, we learned a posterior $P(Z_Q|x)$ by training the VQ-VAE with spectrograms $x\in X_{train}$, whose parametric model is expressed by Eqn.~\eqnref{vqvaefwd}. 

{\bf Learning the $Z_Q$ prior.} To sample from our codebook, allowing generation of 
new (fake) samples, we must learn the prior $P(Z_Q)$ of our discrete latent sequences generated from $X_{train}.$

We will learn $P(Z_Q)$ probability model by training the transformer to learn its autoregressive form $P(Z^i_Q|Z^{1\cdots (i-1)}_Q)$. 
 
After applying $\Phi$ based on a trained codebook of $VQVAE^{case 1}$ to an input $x$, we 
get a sequence  of integers 
$Z^{case 1}_Q(x)=\set{w_1^1, \cdots, w_1^{32\times44}}.$ The same $x$ processed by $VQVAE^{case 2}$  results in $Z^{case 2}_Q(x)=\set{w_2^1, \cdots, w_2^{16\times22}}.$
We generate such sequences for each $x\in X_{train},$ resulting in 2 datasets $DT_1$ and $DT_2$ for training 2 versions of the transformer. It should be noted that $Z^{case(m)}_Q,\ m=1,2$ is a 
representation of $\hat{x}$ obtained using the encoder of $VQVAE^{case(m)},$   which generates it
from $z^{case(m)}_Q(x),$ the non-ordinal version of $Z^{case(m)}_Q(x).$  Thus we simplify 
our problem of generating deep fakes from generative models trained directly on 
spectrograms of $5,632$ real valued 
features  by training generative models $T_1$ and $T_2$ in the discrete latent space using $DT_1$ and $DT_2$
of dimension $n=1,408$ and $n=352,$ respectively.
 
Once we learn the prior $P(Z_Q)$ as a joint autoregressive distribution $P\paren{Z^n_Q|, Z^{n-1}_Q\cdots, Z^{1}_Q}$ of ordered tokens $Z^i_Q,$  after training on the datasets $DT_1$ and $DT_2$, we can feed 
a randomly chosen value for $Z_Q^1 \in \curlb{1,\cdots,|Q|}$ into the respective transformer model and 
have the model predict the next $n-1$ 
values, with $n=1,408$ and $n=352,$ respectively. We then perform the procedure 
explained above to construct our 
spectrogram from the generated indices.

{\bf Autoregressive Models.}  
Our choice for the autoregressive
model is a decoder-only transformer \cite{deconly} (see also Subsection~\ref{subsec:donly}). More 
specifically, we modify an open-source
implementation \cite{monai} of the decoder-only-transformer. Every transformer  
model uses a process called multi-headed self-attention to learn the 
context and relationships between the  
tokens which are the comprising elements of  training sequences.  
The hyperparameters of the VQalAttent transformer are the 
vocabulary size, which is the amount of 
unique input tokens, 
equal to $\abs{Q},$ and the context size $n,$ which 
represents the number of tokens in the latent sequence that the model is able to 
process at a time. For two different values of $n,$ based on two different VQ-VAE compression ratios, we train two different transformers $T_1$ and $T_2.$
In addition to $T_1$ and $T_2,$ referred to as unconditioned transformers, we also train $T^C_1$ and $T^C_2$ by modifying  datasets $DT_1$ and $DT_2,$ so that the sequences of tokens are extended by the class token. The class token is obtained from the label of the original $x\in X_{train},$ used to produce matching sequences $Z_Q(x).$ This is an innovative way of introducing the extrinsic context in the decoder-only transformer, without creating an embedding for it. Sampling the trained $T^C_1$ and $T^C_2$ allows us to obtain deep fakes for which we know the label (digit $c\in\curlb{0,\cdots 9}$). This will be used to assess the quality of the fakes by evaluating the accuracy of their classification on a pretrained classifier (Fig.~\ref{fig:classarch}).
We next analyze the performance of VQalAttent as a function of $n$ and selected hyperparameters.
\section{Evaluation}\label{sec:eval}
Establishing a quantitative metric  for evaluating the effectiveness of a generative model at generating deepfakes is a topic being actively discussed and researched.  In this section, we discuss some of the metrics applied to evaluate both the unconditionally and the conditionally trained generative models. Some methods only  work  for the conditionally generated samples since we know the class of the fakes being generated, based on the class token.

After training the conditioned transformer $T^C_{m}$ for each label $c\in C,\ C=\curlb{0,\cdots, 9}$ equally, we created one dataset of fakes per label $F_c=\curlb{\hat{Z}_{Q_i}^c}_{i=1}^500$ by sampling the model autoregressively, starting from the label token $c \in C.$  Each set $F_c$ represents learned conditional prior distribution $P(Z_Q|c)$. We put all those fakes together into one conditional dataset $F^{(C)}=\curlb{F_c}_{c=0}^9, $ and create another dataset of original discrete latents of the same size and label distribution. We will use these datasets to evaluate the quality of fake latents generated by the conditioned transformers. 

By autoregressively sampling the two unconditioned models $T_1$ and $T_2$ starting from the uninformative token $BOS$ only, we create two datasets of unconditioned fakes $F^{(U)}=\curlb{\hat{Z}_{Q_i}}_{i=1}^5000$ of the same size as $F^{(C)}$.
\subsection{Classifier}
\begin{figure}[h]
\vspace{-2mm}
\centering
\hspace{-1mm}\includegraphics[width=0.46\textwidth, height = 4.2cm]{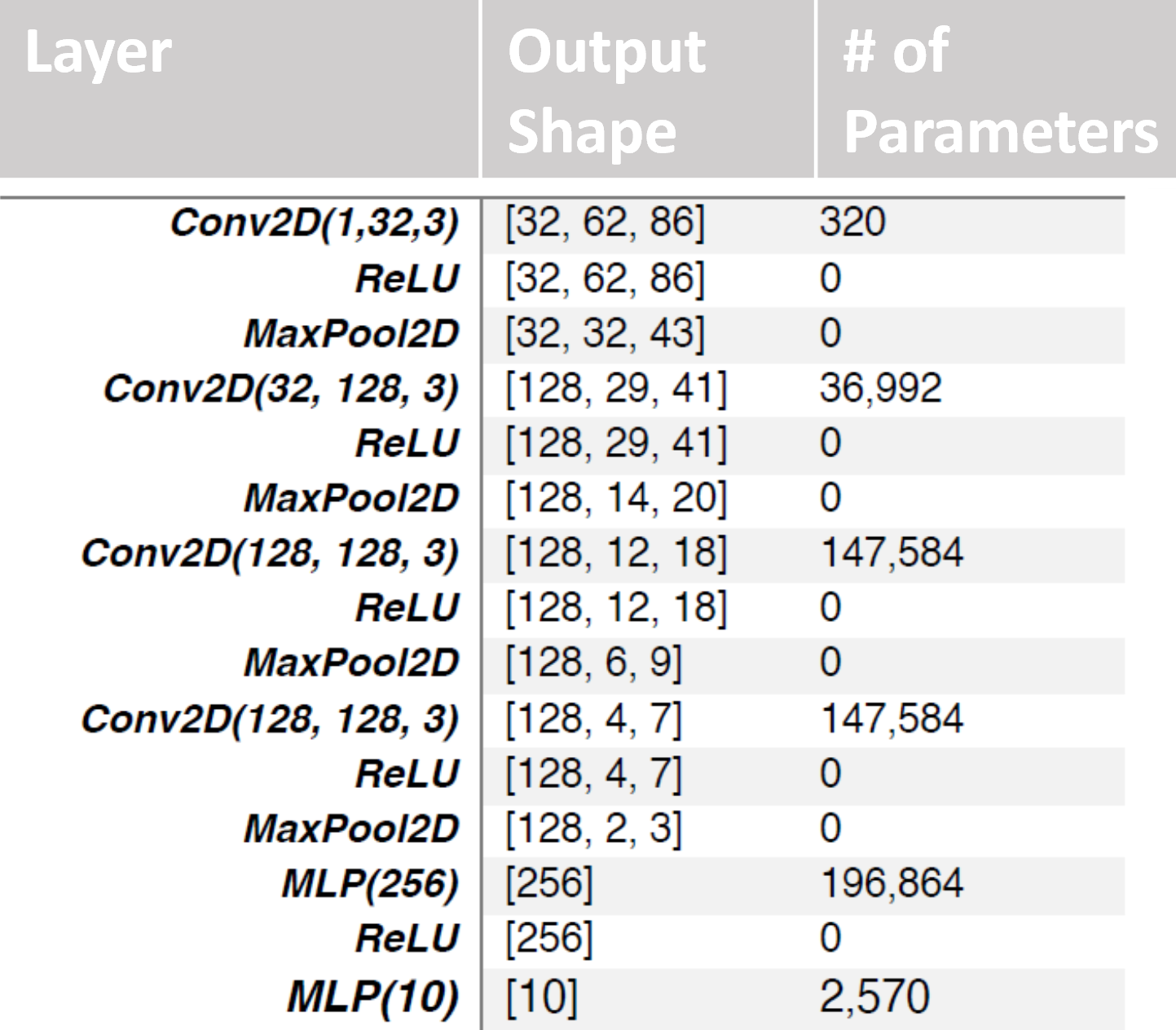}
\vspace{-2mm}
\caption{Architecture of the classifier. Convolutional layers are specified by the 
following format: 
Conv2D(input shape, output shape, stride), fully connected layers are specified by the 
following format:
MLP(number of neurons).
}\vspace{-3mm}
\label{fig:classarch}
\end{figure}

To score the perceptual clarity of the VQ-VAE 
reconstructions in the absence of human judges, and to  quantify the success 
of our conditional generative models in generating the perceptible fakes,  we evaluate the accuracy score achieved by a deep classifier on the reconstructions and 
on the conditionally generated data derived from $F^{(C)}$, respectively. Not having the ground truth label for the fakes generated by the unconditioned transformer, we omit this type of evaluation for $F^{(U)}.$ The classifier 
was trained on the original data $X_{train}$
preprocessed as described. The 
architecture used is of our own design 
and was tuned to give maximum 
accuracy on the test set of original data $X_{test}$ (100\%). The classifier's architecture is described in Fig.~\ref{fig:classarch}.

The classifier-based evaluation outputs the accuracy of the VQ-VAE reconstructions per compression case (1 for lower and 2 for higher compression), and for the respective fakes. It is presented in Table~\ref{tab:acc}, where VQ-VAE reconstructions are marked with letter $R,$ and the spectrograms generated from  the latent fakes $F^{(C)}_u$ produced by the transformer are marked by $S^{(C)}_u,$ where $u=1,2$ for case 1 and 2, respectively.
\begin{table}[h!]
\centering
\vspace{1mm}
\begin{tabular}{ c|c|c|c|c } 
 \hline \\
  & R case 1 & R case 2 & $S^{(C)}_1$ & $S^{(C)}_2$\\
 \hline\\
 Accuracy & 0.966 & 0.961 & 0.912 & 0.909\\ 
 \hline\\
\end{tabular}
\caption{Accuracy scores} \label{tab:acc}
\vspace{-3 mm}
\end{table}
All the accuracy results in Table~\ref{tab:acc} are lower than the accuracy obtained for the original spectrograms, which suggests that the pipeline should have been trained for longer time. We currently train each VQ-VAE model for 100 epochs and each transformer for 50 epochs, which is a strain on our computational resources, given the number of model variants. However, the differences among tested datasets are logical, in that lower compression performs better than the higher compression. 
\subsection{Quantitative evaluation using fidelity and diversity}
A generative model should try to achieve good  trade-off between {\em 1. fidelity} (how realistic each fake is) and {\em 2. diversity} (how well fake samples capture the variations in real samples). We here present the evaluation of both the conditioned and unconditioned generations using the metrics which capture those two aspects in a statistically consistent manner. Fidelity is related to the precision measure, which expresses probability that a random fake falls into the support of the real datapoints, 
while the diversity resembles the recall measure, a probability that a random real sample falls into the support of the fakes \cite{precrecall}. The key word here is the support. The evaluation based on fidelity and diversity defines the support as based only on the statistically relevant samples.

The Topological Precision and Recall (TopP\&R, pronounced 'topper'), \cite{topper} provides a systematic approach to estimating supports of the original and generated data, retaining only topologically and statistically important features with a certain {\em level of confidence}. This allows TopP\&R to stay relevant for noisy features, and provides statistical consistency. Hence, the widespread metrics of precision and recall, applied on the topologically significant features, are called fidelity  and diversity, and used to express relevant qualities in the context of generative sampling. 

Diversity (Recall) counts significant real  samples on the fake manifold, expressing the ability to find all relevant instances of every class and variety in the learned data distribution. Fidelity (Precision) counts significant fake samples on the real manifold, expressing the typicality of the fakes. High number of "hallucinations" would decrease this score. Confidence band for significant samples and the manifolds are estimated using Kernel Density 
Estimator (KDE)  \cite{topper}. Let us denote the relevant fake instances from a randomly generated set of fakes $\curlb{F}$ by $\curlb{F_R},$ indicating that these instances fall withing the real manifold. Similarly, the set $\curlb{R_F},$ collects random real instances from the test set $\curlb{R}$ which fall withing the fake manifold.

${\text{Fidelity}}=\frac{\curlb{F_R}}{\curlb{F}}\\$

${\text{Diversity}}=\frac{\curlb{R_F}}{\curlb{R}}\\$

 Looking at the TopP\&R results in Table~\ref{tab:toppr}, we observe that the diversity is outstanding  across the evaluated pipelines. As we evaluated the model with the fastest training first, we encountered the TopP\&R results from the conditioned transformer $T^C_2$ first: fidelity: 0.904, diversity: 0.986, Top F1: 0.943. We suspected that the diversity is outstanding because we trained conditioned transformers to learn conditional joint distributions of tokens $P(Z^{case 2}_Q|c)$ for each label $c$ separately, and then sampled each of those distributions by starting from the class token. In other words, we obtained good diversity because we forced it by sampling an equal number of samples from each class token. 
Next, we evaluated the matching unconditioned transformer $T_2$, which learned the unconditioned prior $P(Z^{case 2}_Q).$ Using the fakes generated by $T_2$, we obtained somewhat surprising  TopP\&R results (table~\ref{tab:toppr}): the diversity stayed outstanding, but the fidelity went down. By listening to generated fakes, we identified every single digit class and a good variety of male and female voices, which explains high diversity. However, some of the digits were unintelligible due to the noise induced by inverting Mel spectrogram. The fact that some fakes can be clearly identified by human listeners despite the loss due to inversion, indicates that there are good quality fakes that are fairly resilient to the inversion loss. However, those that are unintelligible probably contributed to the drop in fidelity. 
For the $T_1$ and $T^C_1,$ the fidelity further decreased, which we assumed is the consequence of the artifacts due to insufficient training, given that longer token sequences require longer training. In fact, the current Top P\%R values for $T_1$ and $T^C_1$ are obtained by increasing the training to 70 epochs, since the 50 epochs resulted in fidelity close to zero.
Please see the table~\ref{tab:toppr} which displays unconditioned and conditioned TopP\&R values together.  
\begin{table}[h!]
\centering
\begin{tabular}{ c|c|c|c|c } 
& $T_1$ & $T^C_1$  & $T_2$ & $T^C_2$\\
 \hline
 Fidelity  & 0.51 & 0.45 & 0.57 & 0.90\\ 
 Diversity & 1.0 & 1.0 & 1.0 &  0.99\\
 TopF1     & 0.68 & 0.62 & 0.72 & 0.94\\ 
 \hline  
\end{tabular}
\caption{TopP\&R scores for VQalAttent} \label{tab:toppr}
\vspace{-3mm}
\end{table}
We also hypothesize that training the unconditioned
transformers longer, or with a more complex transformer architecture, can decrease the gap in Table~\ref {tab:toppr}. This is because the embedding of the class label in the training sequence decreases the uncertainty in the training data, allowing the transformer model to be easily trained, while the opposite is true for unconditioned sequences of tokens. 
\begin{figure}[h]
\vspace{-0.5mm}
\centering
\hspace{-1mm}\includegraphics[width=0.44\textwidth, height = 3.7cm]{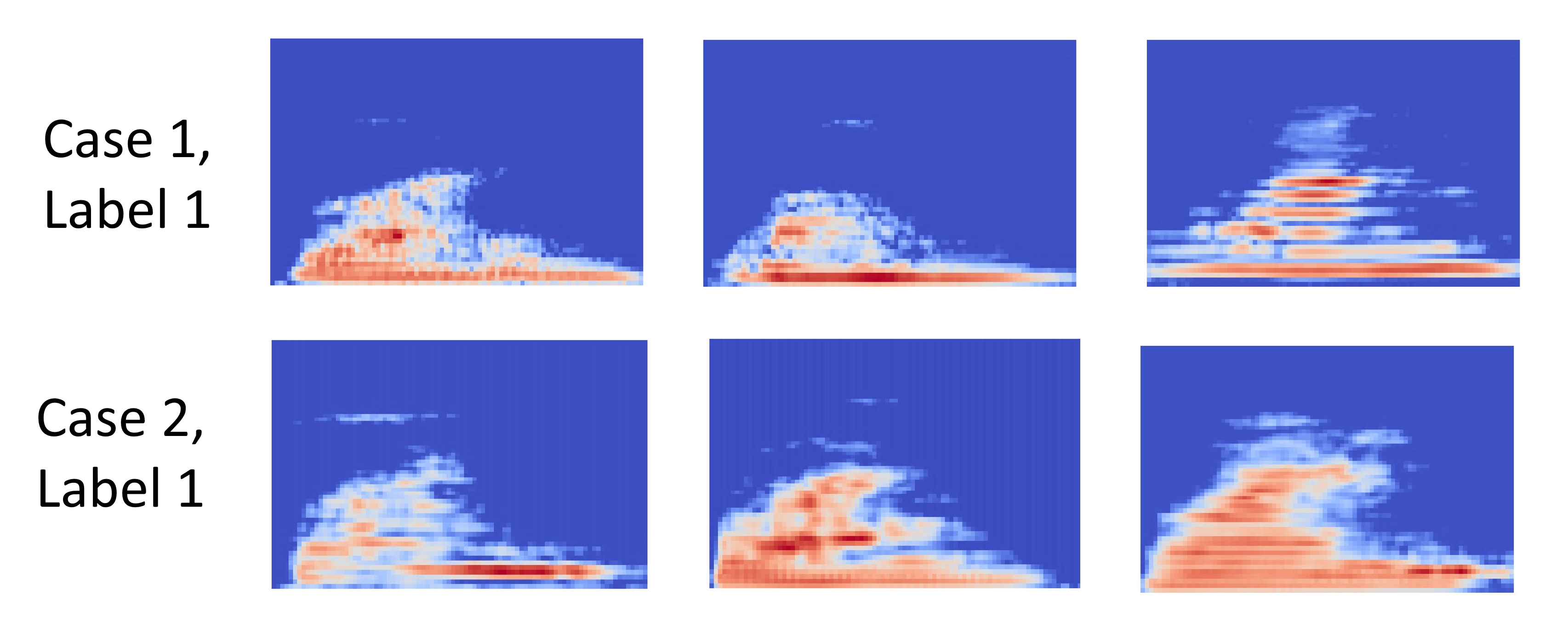}
\vspace{-2mm}
\caption{Deep fakes created by $T^C_{1,2}$ for the class of $1$.
}\vspace{-4mm}
\label{fig:fakeassess}
\end{figure}
\subsection{Qualitative}
\begin{figure}[h]
\vspace{-2mm}
\centering
\hspace{-1mm}\includegraphics[width=0.46\textwidth, height = 4.0cm]{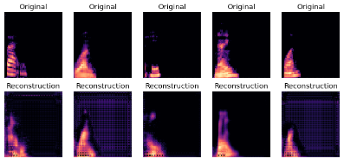}
\vspace{-2mm}
\caption{Original and VQ-VAE reconstructed 
spectrograms.
}\vspace{-3mm}
\label{fig:orig_rec_spec}
\end{figure}

Fig.~\ref{fig:fakeassess} demonstrates how difficult it is to evaluate quality of the fakes, even if they are generated conditionally, for a single class. This is due to the diversity of genders and accents present in the training data. Hence, if properly trained, the transformer will generate very diverse voices, both male and female, and with different accents and annunciation, even if the numeral class is constrained. Inspecting only a few spectrograms (Fig.~\ref{fig:fakeassess}), or playing back a few waveforms derived from such spectrograms is only a practical tool, which can be used to track improvements in the training based on the subjective sense of clarity. Visually comparing originals with VQ-VAE reconstructions in Fig.~\ref{fig:orig_rec_spec} and even with conditionally generated spectrograms can demonstrate some artifacts due to compression, but any  probability shift in the generated fakes (learned distribution) is impossible to detect via visual inspection.
\begin{figure}[h]
\vspace{-2mm}
\centering
\hspace{-1mm}\includegraphics[width=0.46\textwidth, height = 4.0cm]{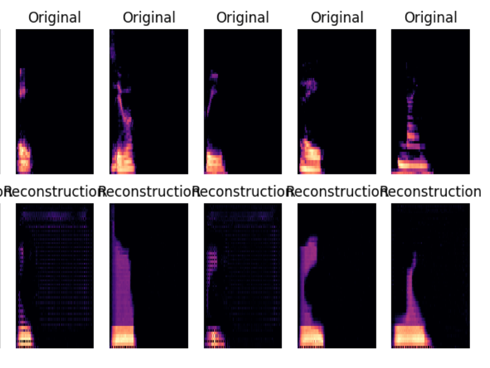}
\vspace{-2mm}
\caption{V(o)QalAttent generated 
spectrograms.
}\vspace{-3mm}
\label{fig:gpt_spec}
\end{figure}

\section{Conclusion}
We  analyzed a two-stage pipeline of a generative model trained to produce deep fakes of spoken numerals.  The pipeline consists of {\bf 1)} a VQ-VAE which discretizes  latent representations of the spectrograms obtained from the AudioMNIST dataset, and {\bf 2)} a decoder-only transformer which is trained to learn the prior of discretized latents. We trained 2 versions of 2 differently dimensioned pipelines,  one version conditioned on the class label matching the spoken numeral's latent, and another where the label was not presented in the training data. Dimensioning of the VQ-VAE/ transformer pipeline was determined by the desired dimension of the latent tensor $z,$ explicitly related to the information reduction ratio. We evaluated all mentioned generative models using diverse metrics, including fidelity, diversity and class recognition, and analyzed the impacts of compression, discretization in the latent space,  presence of the label-based context and model architectures. Future works includes a more rigorous analysis of the architectural effects, additional training,  and the inclusion of diverse data and data transforms in the original and latent space. It will be informed by the dependencies revealed by the VQalAttent analysis, thanks to the simplicity, transparency and modularity of our approach. 
\vspace{-2mm}
\bibliographystyle{IEEEtran}
\bibliography{audiovqatranbib}
\end{document}